\documentclass[letterpaper, 10 pt, conference]{ieeeconf} 

 \IEEEoverridecommandlockouts                              % This command is only needed if 
\usepackage{cite}
\usepackage{orcidlink}

\usepackage{graphicx}
\usepackage{pstricks}
\usepackage{verbatim}
\usepackage{psfrag}
\usepackage[english]{babel}
 \usepackage{amsmath}
 \usepackage{amssymb}
 \usepackage[font=small]{caption}
  \usepackage{fancybox}
 \usepackage{steinmetz}
\usepackage{subcaption}
\usepackage{bodegraph}
\usepackage{gensymb}
\usepackage{algorithm,algorithmic}
\usepackage{amsfonts}
\usepackage{dsfont}
\usepackage{siunitx}
\usepackage{lscape}
\usepackage{enumerate}
\usepackage{lmodern}
\usepackage[T1]{fontenc}
\usepackage[framed]{matlab-prettifier}
\usepackage{mathtools}
\usepackage{color} %red, green, blue, yellow, cyan, magenta, black, white
\usepackage{xcolor,colortbl}
\usepackage{colortbl}
\usepackage{multirow, hhline}
\usepackage{url}
\usepackage{hyperref}
\usepackage{multicol}
\usepackage{supertabular}

\definecolor{verdemio}{rgb}{0.01, 0.75, 0.24}

\renewcommand{\keywords}[1]{\textbf{\textit{Index terms---}} #1}

\makeatletter
\newcommand\notsotiny{\@setfontsize\notsotiny\@vipt\@viipt}
\makeatother

\graphicspath{{Images/}}

\definecolor{Gray}{gray}{0.9}

\definecolor{verdedd}{rgb}{0.0, 0.5, 0.0}

 \newtheorem{Optimi}{\bf Optimization Problem}

\definecolor{myorangenew}{rgb}{1 0.49 0}
\definecolor{mygreennew}{rgb}{0.31 0.78 0.47}
\definecolor{mygreynew}{rgb}{0.6 0.6 0.6}
\definecolor{bluD}{rgb}{0.09, 0.45, 0.81}
\definecolor{bluD_l}{rgb}{0.29, 0.65, 1}
\definecolor{nero}{rgb}{0, 0, 0}
\definecolor{bianco}{rgb}{1, 1, 1}

\definecolor{bluG}{cmyk}{100 85 0 0}
\definecolor{verdeG}{cmyk}{75 0 100 0}

\def\e{{\boldsymbol e}}

 \def\r{{\boldsymbol r}} \def\s{{\boldsymbol s}} \def\t{{\boldsymbol t}}

 \def\x{{\boldsymbol x}}
\def\y{{\boldsymbol y}}

\newtheorem{Remar}{\bf Remark}

\DeclareMathSymbol{\boldmu}{\mathord}{letters}{15}
\title{\LARGE \bf 
Arc-Length-Based Warping for Robot Skill Synthesis from Multiple Demonstrations}
\author{Giovanni Braglia$^{1}$\orcidlink{0000-0002-2230-8191}, Davide Tebaldi$^{2}$\orcidlink{0000-0003-1432-0489}, André E. Lazzaretti$^{3}$\orcidlink{0000-0003-1861-3369}, 
and Luigi Biagiotti$^{2}$\orcidlink{0000-0002-2343-6929} %
 \thanks{The work has been partly supported by the University of Modena and Reggio Emilia,
through the action FARD (Finanziamento Ateneo Ricerca Dipartimentale) 2022-2024, and by the ECOSISTER Project. Project funded under the National Recovery and Resilience Plan (NRRP), Mission 04 Component 2 Investment 1.5 – NextGenerationEU. Call for tender n. 3277 dated 30/12/2021. Award Number:  0001052 dated 23/06/2022.}%
 \thanks{$^{1}$The author is with the Italian Institute of Technology, Genova, Italy
 {\tt\small giovanni.braglia@iit.it}.%}
$^{2}$The authors are with the University of Modena and Reggio Emilia, Italy
{\tt\small \{davide.tebaldi, luigi.biagiotti\}@unimore.it}.
$^{3}$ The author is with the Federal Technological University of Paraná (UTFPR), Curitiba, Brazil. {\tt\small lazzaretti@utfpr.edu.br} }%
\thanks{\noindent AutoLab Co-Manipulation Dataset:   
\textit{ https://github.com/AutoLabModena/AutoLab-Co-Manipulation-Dataset.git}}
\thanks{\noindent Additional Material: \textit{https://github.com/AutoLabModena/Spatial-Sampling.git}}%
\thanks{\noindent Video: \textit{https://youtu.be/BTxUMeL-ebI}}
}

\begin{document}

\maketitle
\thispagestyle{empty}
\pagestyle{empty}

\begin{abstract}
In robotics, Learning from Demonstration (LfD) aims to transfer skills to robots by using multiple demonstrations of the same task. These demonstrations are recorded and processed to extract a consistent skill representation. This process typically requires temporal alignment through techniques such as Dynamic Time Warping (DTW). In this paper, we consider a novel algorithm, named Spatial Sampling (SS), specifically designed for robot trajectories, that enables time-independent alignment of the trajectories by providing an arc-length parametrization of the signals. This approach eliminates the need for temporal alignment, enhancing the accuracy and robustness of skill representation, especially when recorded movements are subject to intermittent motions or extremely variable speeds, a common characteristic of operations based on kinesthetic teaching, where the operator may encounter difficulties in guiding the end-effector smoothly.
To prove this, we built a custom publicly available dataset of robot recordings to test real-world movements, where the user tracks the same geometric path multiple times, with motion laws that vary greatly and are subject to starting and stopping. The SS demonstrates better performances against state-of-the-art algorithms in terms of (i) trajectory synchronization and (ii) quality of the extracted skill.
\end{abstract}


%%%%%%%%%%%%%%%%%%%%%%%%%%%%%%%%%%%%%%%%%%%%%%%%%%%%%%%%%%%%%%%%%%%%%%%%%%%%%
\section{Introduction}\label{sec:Introduction}
%%%%%%%%%%%%%%%%%%%%%%%%%%%%%%%%%%%%%%%%%%%%%%%%%%%%%%%%%%%%%%%%%%%%%%%%%%%%%

% Robot Learning from Demonstration (LfD) teaches robots tasks via a few demonstrations~\cite{pastor2009learning,calinon2010learning}. Specifically, LfD seeks to facilitate intuitive skill transfer, reduce the need for extensive data, and make robots more approachable to individuals without programming expertise~\cite{pastor2009learning}.

\begin{figure}[t]
    \centering
    \psfrag{a}[t][t][0.7]{(1) Collect}
    \psfrag{e}[t][t][0.7]{demonstrations}
    \psfrag{b}[t][t][0.7]{Signals Alignment}
    \psfrag{c}[t][t][0.7]{Skill Extraction}
    \psfrag{d}[t][t][0.7]{Skill Encoding}
    \psfrag{f}[t][t][0.7]{ (2) Offline Skill Computation}
    \psfrag{g}[t][t][0.7]{Robot}
    \psfrag{h}[t][t][0.7]{Environment}
    \psfrag{j}[t][t][0.7]{(3) Execution}
    \begin{subfigure}{0.95\columnwidth} 
    \setlength{\abovecaptionskip}{-1pt}
    \includegraphics[width=0.84\linewidth]{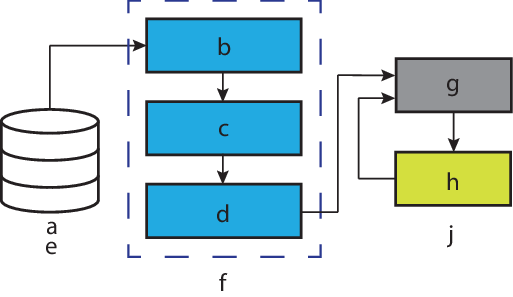}    %trim={4cm 0 3.8cm 3.8cm}, 
       \caption{}
        \label{subfig:LbD_Schematic}
    \end{subfigure}
    \begin{subfigure}{0.95\columnwidth} 
    \setlength{\abovecaptionskip}{-1pt}
        \includegraphics[width=0.94\linewidth]{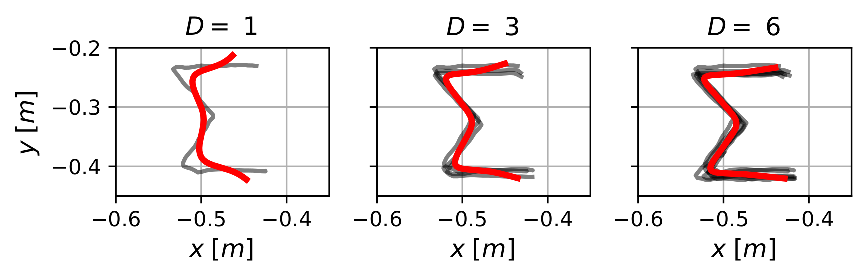} 
        \caption{}
        \label{subfig:demonstrations}
    \end{subfigure}
    \caption{Learning from Demonstration (LfD) framework (a) and approximation accuracy of the computed barycenter (red) with varying number of demonstrations $D$ (black) (b).} 
\vspace{-0.5cm}
\label{fig:LbD_Schematic}
\end{figure}

Robot Learning from Demonstration (LfD) teaches robots tasks via a few demonstrations, aiming to facilitate intuitive programming with no need of prior robot expertise, and reduce the need for extensive data~\cite{pastor2009learning,calinon2010learning}.
LfD architectures generally follow a three-step workflow, as illustrated in Fig.\ref{subfig:LbD_Schematic}: (1) collecting demonstrations, (2) extracting relevant information, commonly referred to as the skill in robotics literature, and (3) executing the learned task while interacting with the environment\cite{gavspar2018skill,bianco2017scaling}.
{\black Generally, to synthesize a single skill, several demonstrations are provided to improve accuracy with respect to the desired behavior. }
For this reason, Step (2) can be further decomposed into: (i) aligning the collected demonstrations, (ii) extracting a representative pattern, referred to as the barycenter in this paper, and (iii) encoding the learned behavior for reproduction. This work focuses on (i) and (ii), proposing a method applicable to any encoding technique.

Barycenter computation typically relies on averaging techniques across demonstrations. Gaussian Mixture Models (GMM) with regression (GMR)\cite{calinon2007learning}, for instance, can estimate a barycenter even from a single demonstration, as shown in Fig.\ref{subfig:demonstrations}-left. However, increasing the number of demonstrations improves accuracy, as illustrated in Fig.~\ref{subfig:demonstrations}-right. Since acquiring numerous demonstrations is not always feasible or convenient, LfD approaches often integrate pre-processing techniques to maximize the information extracted from limited data.
As mentioned above, a critical step involves aligning the collected signals, typically achieved by defining a similarity or variance metric within the demonstration set to extract the skill. In this context, timing frequently introduces distortions, as temporal misalignments can compromise the accuracy of comparing time series data~\cite{su2021toward}.Techniques like DTW measure similarity regardless of timing, but DTW doesn't fully decouple time, which is crucial in many robotics applications~\cite{muller2007dynamic,senin2008dynamic}. For instance,
one may want to adjust robot trajectories afterward to account for geometric, kinematic, or external constraints~\cite{braglia2024phase}.
This paper addresses these two problems, i.e., the time-decoupling and synchronization, often referred as warping, of time series by proposing the so-called Spatial Sampling (SS) algorithm. The SS is tested for robotics applications, carrying additional geometric properties that will be discussed in the following sections. 
The paper is structured as follows: Section II provides a review on the affine literature. Section III introduces the proposed SS algorithm, providing an analytical study on the choice of the sampling period $\Delta$. Section IV evaluates SS's alignment and approximation performance through experiments, and Section V concludes the study.\\[-3mm]

%----------------------------------------------------------------------------
\section{Related Works}\label{sec:Related Works}
%----------------------------------------------------------------------------

% LfD draws inspiration from the human ability to learn and generalize from just a few observations. 
% The goal of LfD is to extract relevant knowledge from limited demonstrations, equipping the robot with the expertise to handle and generalize over future tasks~\cite{pastor2009learning}. Demonstrations can be provided to robots through various methods, such as kinesthetic teaching, teleoperation, videos, or vocal commands, while recording different types of data, including Cartesian/joint trajectories, force measurements, and images, among others~\cite{su2021toward,pignat2022}. 
% In this paper we use kinesthetic teaching, namely hand guiding the robot along the desired task demonstration, with specific focus on workspace trajectories. 

The goal of LfD is to extract relevant knowledge from limited demonstrations, enabling robots to generalize across future tasks~\cite{pastor2009learning}. Demonstrations can be provided through kinesthetic teaching, teleoperation, videos, or vocal commands, capturing data such as Cartesian/joint trajectories, force measurements, images, among others~\cite{su2021toward,pignat2022}. In this work, we employ kinesthetic teaching, guiding the robot manually to record workspace trajectories for task demonstrations.

When extracting the skill, a key step is aligning (or warping) demonstrations to prevent temporal shifts from introducing noise~\cite{calinon2007learning}. In signal processing, Dynamic Time Warping (DTW) is a simple yet robust technique for achieving this~\cite{muller2007dynamic,senin2008dynamic}. In the basic case of comparing two-time series, alignment refers to establishing a point-by-point correspondence between the signals, minimizing a chosen variance metric. For DTW, this metric is the Euclidean distance and it is used to compute a distance matrix that allows tracing back an optimal alignment path between the two time series~\cite{chiu2004content}.

Once all trajectories in the dataset are aligned, consistent skill extraction can be achieved through time series averaging, commonly referred to as consensus representation or, more easily, barycenter computation. Initially, the term barycenter referred to an average sequence that minimizes the squared distance to all series in the demonstration library~\cite{petitjean2011global}. Since then, alternative methods for barycenter computation have been developed. For instance, in~\cite{vayer2020time}, the definition of barycenter was adapted to ensure global invariance to spatial alterations, such as rigid transformations. Alternatively, in~\cite{shapira2019diffeomorphic}, a learning-based approach uses transformer layers to align input signals, then computes the barycenter by simply averaging them.
In robotics, barycenter computation is typically performed using a combination of Dynamic Time Warping (DTW) and Gaussian Mixture Models/Regression (GMM/GMR)~\cite{su2021toward,calinon2007learning,zeng2022gmmdtw}. By first pre-processing the recorded robot trajectories with DTW, time shifts can be removed, allowing the geometric variance to be encoded into the GMM. This provides valuable insights into how the dataset deviates from the regression model (GMR), which can be used to enhance task execution~\cite{pignat2022}.

While efficient and robust, the previously mentioned techniques are inevitably sensitive to time variations, as the retrieved barycenter is tied to a specific duration. Moreover, in certain robotics applications, the primary focus may be on the geometric path constraint, while the execution velocity is determined later~\cite{braglia2024minj,raiola2015co}. The idea is to define a path from the recorded trajectories as a function of its arc-length~\cite{braglia2024phase}. This parameter, referred to as $s$ throughout this paper, defines the direction of travel along the curve and indicates the distance covered~\cite{todorov1998smoothness}. By adjusting the feed rate $\dot{s}$ afterward, the robot's task execution can be modulated, allowing for the integration of internal or external constraints~\cite{bianco2017scaling}.
In~\cite{gavspar2018skill,CIFUENTES2019162}, the authors demonstrate that switching from the time domain to the arc-length parameter domain can yield better classification features for action recognition in the context of robot motion trajectories. In this paper, we propose the use of a Spatial Sampling (SS) algorithm to compute the arc-length parametrization of workspace trajectories directly from the demonstrated motions. Originally studied in~\cite{braglia2024phase}, the SS algorithm filters the input trajectory and generates a path composed of equally spaced points. As discussed later, the SS algorithm conceptually resembles DTW, but it operates in the arc-length domain instead of the time domain. While DTW is alignment-specific, we explore how the SS algorithm can be adapted for alignment purposes and show that skill extrapolation becomes more effective when transitioning to the arc-length domain in cases with multiple demonstrations.
\vspace{-2mm}

%----------------------------------------------------------------------------
\subsection{Contributions}\label{subsec:Contributions}
%----------------------------------------------------------------------------

In this paper, with respect to \cite{braglia2024phase}, (i) we extend the SS algorithm to scenarios in which multiple demonstrations are provided and (ii) provide a methodology to choose its parameter to achieve target approximation accuracy, together with an analysis of its computational cost. Additionally, we study the SS performance in terms of (iii) trajectory alignment and (iv) approximation accuracy, with extensive comparisons with state-of-the-art related algorithms. Finally (v) we release an open-source robot recordings dataset for skill extraction.

\section{Spatial Sampling}\label{sec:Spatial Sampling}
Arc-length parametrization is a fundamental concept in differential geometry \cite{toponogov2006differential} and widely applied in robotics \cite{bianco2017scaling}. Despite its importance, it is difficult to find robust algorithms for computing arc-length from time-based position trajectories. For example, the method proposed in\cite{gavspar2018skill} cannot handles zero-speed intervals. Additionally, low-speed regions can cause dense point distributions, leading to interpolation oscillations known as the Runge effect.
{\black Moreover, it is necessary to stress that demonstrated trajectories are always recorded according to a time-based framework, thus embedding all the temporal information of the execution.}

The proposed Spatial Sampling (SS) algorithm generates the sequence $\x_\Delta$ of points belonging to the new filtered trajectory $\r(s_\Delta)$, 
which is a function of the arc-length parameter $s$, starting from the sequence $\r_{i}=\r(iT_s)$, $i = 0, \ldots, N$, recorded with sampling time $T_s$. 
The proposed algorithm works as follows:

\noindent 1) Given the samples 
 $\r_{i}$, a linearly interpolating continuous-time function $\r_L(t)$ is built. If the sampling period $T_s$ is small enough, then the following holds true:
\begin{equation}\label{eq:new_eqq_2}
    \r_L(t)\approx \r(t).
\end{equation}
\noindent 2) A new sequence  $\x_{\Delta,k}$ is created by imposing that $\x_{\Delta,0} =  \r_L(0)$ and $\x_{\Delta,k} =  \r_L(t_k)$ for $k>0$, where $t_k$ is the time instant that guarantees the following condition:
\begin{equation}\label{eq:norm_Delta1}
\| \x_{\Delta,k}-\x_{\Delta,k-1} \| =\Delta
, \;\;\; \mbox{for} \;\;\; k=1,\ldots,K.
\end{equation}
\begin{figure}[tb]
\centering
    \begin{subfigure}{0.49\columnwidth}    
    \centering
    \includegraphics[width=1\linewidth]{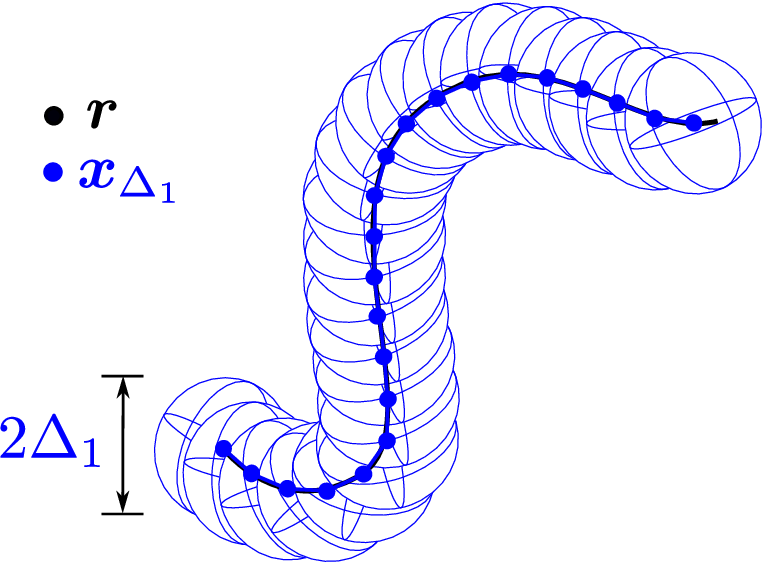}
        \caption{}
        \label{subfig:delta1}
    \end{subfigure}\hspace{-5mm}
    \begin{subfigure}{0.49\columnwidth}
    \centering
        \includegraphics[width=1\linewidth]{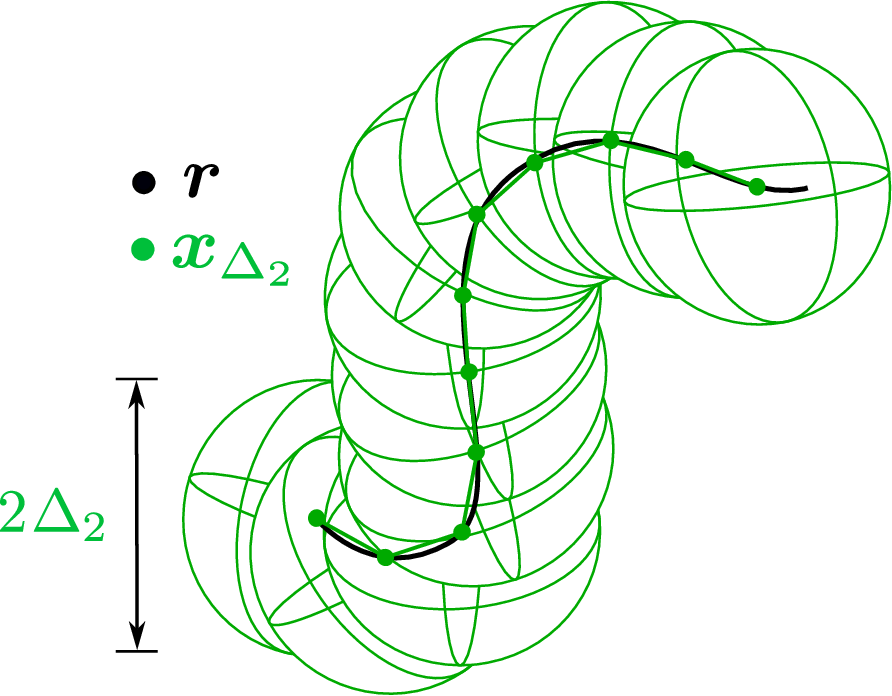}
        \caption{}
        \label{subfig:delta2}
    \end{subfigure}
    \caption{Working principle of the Spatial Sampling (SS) algorithm for two different $\Delta$ values. Lower $\Delta$ induces a better approximation of the recorded curve $\r$. Note that the condition\eqref{eq:norm_Delta1} focuses on a spatial constraint, allowing the SS algorithm to be independent on the demonstration's timing.}
    \label{fig:SpSa_1D_example_05_1}
%\vspace{-0.5cm}
\end{figure}
The parameter $\Delta$ defines the geometric distance between consecutive samples of the filtered trajectory and can be freely chosen.
The condition in~\eqref{eq:norm_Delta1}
implies that the total distance over the curve between the first point $\x_{\Delta,0}$ and the generic $k$-th point $\x_{\Delta,k}$
is given by  $k\Delta$, which approximates the length of the curve $\r_L(t)$ at the time instant  $t_k$ with a precision that increases as $\Delta$ decreases. Therefore, for $\Delta$ sufficiently small, the SS algorithm introduces the following mapping between  the length $s_k = k\Delta$ and the position along the  approximating linear curve $\r_L$:
\begin{equation}\label{eq:Spatial_sequence}
 \x_{\Delta,k}=\r_L(t_k), \;\;\;\mbox{ with }\;\;\; t_k = \gamma^{-1}(s_k),
 \end{equation}
where $s(t) = \gamma(t)$ is the particular timing law imposed during the trajectory demonstration, 
describing how the robot moves along the imposed geometric path since the variable $s$ is the arc-length parameterization of the curve. 
From~\eqref{eq:new_eqq_2},~\eqref{eq:norm_Delta1} and ~\eqref{eq:Spatial_sequence}, the resulting trajectory parameterized with respect to the 
arc-length parameter $\x_\Delta(s)\approx \r_L(s) = \r_L(\gamma^{-1}(s))$ can be obtained. Therefore, the sequence $\x_{\Delta,k}$ is the result of a sampling operation with a constant spatial period $\Delta$, characterized by:
\begin{equation}\label{eq:unitary_norm}
{\left\lVert\dfrac{d \x_\Delta(s)}{d s}\right\rVert}_{s=s_k} \approx
\dfrac{\|\x_{\Delta,k+1}-\x_{\Delta,k}\|}{\|s_{k+1}-s_{k}\|}
=\!\dfrac{\Delta}{\Delta}\!
=1.
\end{equation}

Consequently, the derivative in  \eqref{eq:unitary_norm} will always be different from zero, and the tangential direction of the curve will always be well-defined. The operation of the spatial sampling algorithm is graphically shown in Fig.~\ref{fig:SpSa_1D_example_05_1}. The pseudo-code is shown Algorithm~\ref{algorithm:SS}; while the python and MATLAB %version of the SS is then 
codes are provided in the additional material.
\vspace{-2mm}
% The operation of the spatial sampling algorithm is graphically shown in the case study of Fig.~\ref{fig:SpSa_1D_example_05_1}. In  Fig.~\ref{fig:SpSa_1D_example_05_1}(a) the approximating linear function $\r_L(t)$ is shown, from which it can be seen that the original geometric distance between two consecutive samples $y_{i}$
% is not constant. Subsequently, in Fig.~\ref{fig:SpSa_1D_example_05_1}(b), the spatial sampling algorithm is applied to the
% curve $\ylscal(t)$ using a certain spatial interval $\Delta=\tilde\Delta$,
% where it can be noticed that a more accurate approximation of the original curve can be achieved for smaller parameters $\Delta$. Finally, Fig.~\ref{fig:SpSa_1D_example_05_1}(c) shows the samples $s_k$ of the function $s = \gamma(t)$ introduced in \eqref{eq:Spatial_sequence}. Fig.~\ref{fig:SpSa_1D_example_05_1}(a) shows that the proposed spatial sampling imposes no constraints on the demonstrated trajectory $\y(t)$, which can also include parts with zero speed, which might occur if the user stops during the demonstration. This is an important feature since no segmentation of the demonstrated trajectory is required, as it is instead by alternative approaches~\cite{gavspar2018skill,cifuentes2013arc}.
%
%
%
\begin{algorithm}[t] 
 \caption{Spatial Sampling (SS)}
 \begin{algorithmic}[1]
 \renewcommand{\algorithmicrequire}{\textbf{Input: $\r \in \mathds{R}^{(N\times d)}, 
 \; \t \in \mathds{R}^{(N\times 1)}, 
 \; \Delta$}} 
 \renewcommand{\algorithmicensure}{\textbf{Output: $ \t_\Delta \in \mathds{R}^{(K\times 1)},
 \; \x_\Delta \in \mathds{R}^{(K\times d)},
 \; \s_\Delta \in \mathds{R}^{(K\times 1)}$} $\hspace{2cm}$}
 \REQUIRE 
 \ENSURE  
 \textit{Initialization}: \\ 
 \STATE Initialize arrays: $\t_{\Delta,0} \!= \!0$, $\x_{\Delta,0} \!= \!\r_0$, $\s_{\Delta,0} \!= \!0$
 \STATE Initialize current state: $\y_c \!= \! \r_0$, $i=0$
 \\ \textit{Compute $ \t_\Delta, \; \x_\Delta, \; \s_\Delta$} : \\
 \WHILE{$i<N$}
  \IF{ $|| \r_{i+1} - \y_c|| > \Delta$}
  \STATE Compute $\x_k$ belonging to the segment $ \overline{\r_i\r_{i+1}} $ and such that $|| \x_{k} - \y_c|| = \Delta $
  \STATE Compute $t_k$ to ensure~\eqref{eq:Spatial_sequence}
  \STATE Update arrays: $\t_\Delta \! = \! [\t_\Delta, t_k]$, $\x_\Delta \!=\! [\x_\Delta, \x_k]$, $\s_\Delta \!=\! [\s_\Delta, \s_\Delta(end)+\Delta]$
  \STATE Update current state: $\y_c \!=\! \x_\Delta(end)$
  \ELSE
  \STATE $i=i+1$
  \ENDIF
 \ENDWHILE
  % \STATE Evaluate $\m(s_{\bar{t}+1}^*)$
 \RETURN $ \t_\Delta, \; \x_\Delta, \; \s_\Delta$
 \end{algorithmic} 
\label{algorithm:SS}
\end{algorithm}
\begin{Remar}\label{rem:curve_regular_remark}
The SS algorithm generates a filtered curve $\x_\Delta(s)$ satisfying condition~\eqref{eq:unitary_norm}. As a consequence, the regularity condition $d \hat \x_\Delta(s)/d s \neq \boldsymbol{0} \quad \forall s \in [0, s_{max}]$ is met~\cite{toponogov2006differential}. This property guarantees that the parametrization $\x_\Delta^\prime(s)$ remains well-defined for every $s$.
\end{Remar}
If replicating the demonstrated reference $\r_t$, $t \in [0, T]$, is required, an important implication of Remark~\ref{rem:curve_regular_remark} is that this problem reduces to finding $\dot{s}_t^\star$ such that $\dot{\x}(s_t^\star) = \dot{\r}_t$, given the relationship $\dot{\x}(s_t) = \x^\prime(s_t) \dot{s}_t$. Specifically, this condition leads to
\[
\dot{\x}(s_t) = \x^\prime(s_t) \dot{s}_t = \dot{\r}_t \implies \dot{s}_t^\star \triangleq \big(\x^{\prime}(\gamma_t)\big)^{-1} \cdot \dot{\r}_t, 
\]
where the existence of $\big(\x^{\prime}(\gamma_t)\big)^{-1}$ is guaranteed as a direct consequence of the curve's regularity.
\vspace{-3mm}

%----------------------------------------------------------------------------
\subsection{Optimization of Parameter $\Delta$}\label{subsec:optimization}
%----------------------------------------------------------------------------
The function $\x_\Delta(s)$ is an approximation of the original trajectory that depends on parameter $\Delta$.
The error between the original trajectory $\r(t)$ and the filtered trajectory $\x_\Delta(s)$ can be evaluated using the well-known Hausdorff distance $d_H$
% ~\cite{Donoso2008}
 as a metric:
\begin{equation}\label{Hausdorff_Distance}    
d_H=\max{\left\{\sup_{\r_{i} \in S_{\r}}d(\r_{i},S_{\x_\Delta}), \sup_{\x_{\Delta,k} \in S_{\x_\Delta} }d(S_{\r},\x_{\Delta,k})\right\}},
\end{equation}
where $S_{\x_\Delta}$ denotes the set of all the samples $\x_{\Delta,k}$ of the filtered trajectory $\x_\Delta(s)$, $S_{\r}$ denotes the set of all the samples $\r_{i}$ of the original trajectory $\r(t)$, and
\[
\begin{array}{c}
d(\r_{i},S_{\x_\Delta})=\inf_{\x_{\Delta,k} \in S_{\x_\Delta}}d(\r_{i},\x_{\Delta,k}),\\[3mm]
d(S_{\r},\x_{\Delta,k})=\inf_{\r_{i} \in S_{\r}}d(\r_{i},\x_{\Delta,k}).
\end{array}
\]
The choice of parameter $\Delta$ in the SS algorithm directly affects $d_H$ in \eqref{Hausdorff_Distance}:
$d_H=d_H(\Delta)$. Therefore, it is of interest to set up the following optimization problem.
\begin{Optimi}\label{opt:optim_delta}
Let $d_H^*$ be the maximum admissible value for the Hausdorff distance, and let $d_{H,t}$ be the acceptable lower tolerance boundary for $d_H^*$.
% , as depicted in Fig.~\ref{fig:Max_Error_Tolerance}.
The objective is to find the optimal value $\overline{\Delta}$ such that: 
\begin{equation}\label{eq:optim_probl_1}
\begin{array}{@{\!\!}c}
\displaystyle \min_{\Delta} J(\Delta), \; %\hspace{8mm}\mbox{where}\; \\[6mm] 
J(\Delta)\!=\!\left\{
\begin{array}{l@{\hspace{1.68mm}}l}
\small
\e_{d_H}(\Delta) & \mbox{if} \hspace{1.68mm} 0< \Delta_{d_H}(\Delta)\! \le\! d_{H,t}, \\[2mm]
\infty & \mbox{otherwise}, 
\end{array}
\right.
\end{array}\!\!\!\!\!
\end{equation}
where $\e_{d_H}(\Delta)=d_H^*\!-\!d_H(\Delta)$.The optimal value $\overline{\Delta}$ solving ~\eqref{eq:optim_probl_1} is the maximum value of parameter $\Delta$ such that $d_H(\overline{\Delta})$ approaches $d_H^*$ from the left.
 \end{Optimi}
%

%............................................................................
\subsubsection{Numerical case study}\label{subsubsec:Numerical example}
%............................................................................
%
 \begin{figure}[t]
  \centering
 \psfrag{y [m]}[l][l][0.8]{$y$ [m]}
\psfrag{x [m]}[b][b][0.8]{$x$ [m]}
\psfrag{ykkkk}[][][0.62]{$\;\;\;\r_L(t)$}
\psfrag{yLkkkk}[][][0.62]{$\;\,\r_L(t_k)$}
\psfrag{yLskkkk}[b][b][0.62]{$\x_\Delta(\s_k)$}
\psfrag{Emmm}[][][0.62]{$d_H(\Delta^\star)$}
 \psfrag{Em}[][t][0.8]{$d_H(\Delta)$}
\psfrag{delta}[][][0.8]{$\Delta$ [log scale]}
\psfrag{Ems}[b][b][0.68]{$\;\;\;\;E_m(\Delta^\star)$}
\psfrag{ds}[b][b][0.68]{$\Delta^\star$}
  \includegraphics[clip,width=\columnwidth]{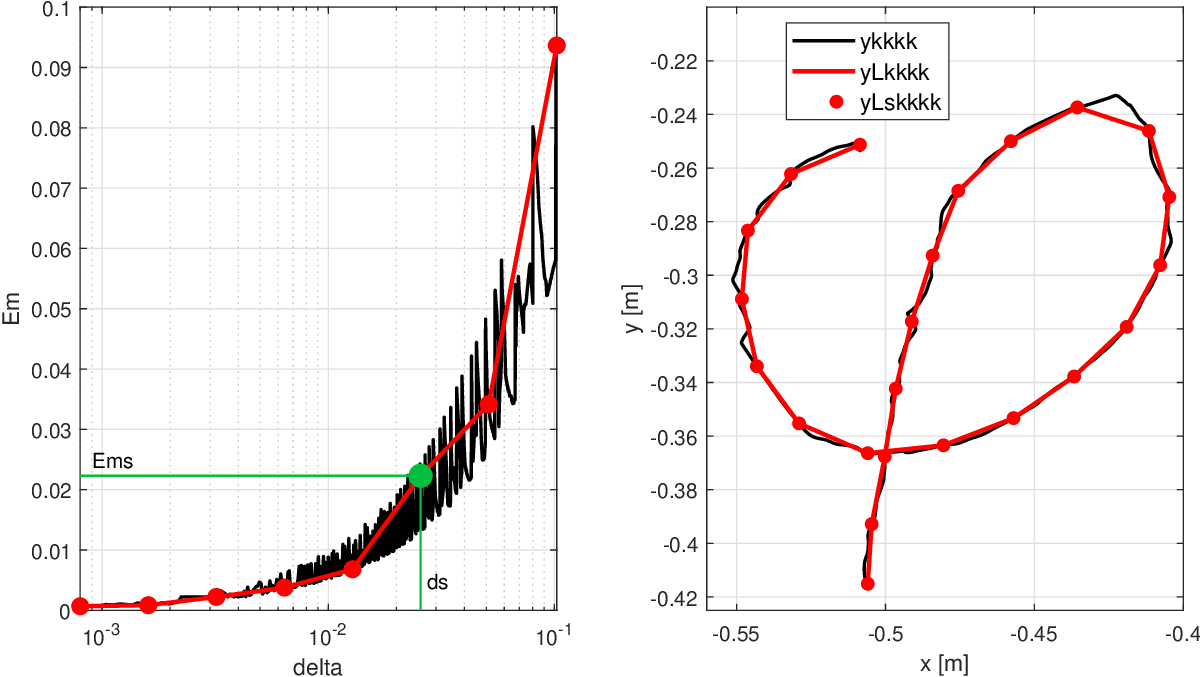}
%       \setlength{\unitlength}{5.0mm}
% \psset{unit=\unitlength}
%  \rput(-7.132,10.05){\footnotesize (a)}
%  \rput(2.168,10.05){\footnotesize (b)}
  %
{\small (a) \hspace{40mm} (b)}
 \caption{Numerical example of Sec.~\ref{subsubsec:Numerical example}. (a) Result of the Optimization Problem~\ref{opt:optim_delta} applied to the trajectory shown in (b) using $d_H^*=0.025$ and $d_{H,t}=0.1\cdot 10^{-3}$. (b) Linear interpolation $\r_L(t)$ of the samples of the original trajectory, samples $\x_\Delta(\s_k)$ of the filtered trajectory using the optimal $\Delta^\star=0.0256$, and their linear interpolation $\r_L(t_k)$.}
\label{Spatial_Sampling_Optim_Example_NEW}
\vspace{-0.4cm}
\end{figure} 
 A numerical case study has been implemented using the Optimization Problem~\ref{opt:optim_delta}, by setting $d_H^*=0.025$ and $d_{H,t}=0.1\cdot 10^{-3}$, and using the original trajectory described by the linear interpolation $\y_L(t)$ shown in Fig.~\ref{Spatial_Sampling_Optim_Example_NEW}(b). The black characteristic of Fig.~\ref{Spatial_Sampling_Optim_Example_NEW}(a) shows the evolution of function $d_H(\Delta)$ as a function of $\Delta$, showing a non-monotonic behavior. The red characteristic shows the evolution of function $d_H(\Delta)$ when $\Delta$ takes on values from the following progression: 
 \begin{equation}\label{geom_prog}
\Delta \in [\Delta_1,\,2\Delta_1,\,4\Delta_1,\,\ldots],
 \end{equation} 
 where $\Delta_1$ is a proper initial condition, showing a monotonic behavior in this case.   By using the geometric progression \eqref{geom_prog}, the Optimization Problem~\ref{opt:optim_delta} is therefore well-defined and has a unique solution. 
 The resulting $\Delta^\star$ for the considered numerical case study is $0.0256$, for which the corresponding $d_H(\Delta^\star)$ is $0.0223$, as shown in Fig.~\ref{Spatial_Sampling_Optim_Example_NEW}(a).  
 % as shown in Fig.~\ref{fig:Spatial_Sampling_Optim_Example_bis}. 
 The linear interpolation $\y_L(t)$ of the original %A-shaped
 trajectory, the samples $\hat{\y}(\s_k)$ of the filtered trajectory using the optimal $\Delta^\star=0.0256$, and their linear interpolation $\y_L(t_k)$ are shown in Fig.~\ref{Spatial_Sampling_Optim_Example_NEW}(b).

%%%%%%%%%%%%%%%%%%%%%%%%%%%%%%%%%%%%%%%%%%%%%%%%%%%%%%%%%%%%%%%%%%%%%%%%%%%%%
\section{Experiments and Evaluation}\label{sec:Experiments and Evaluation}
%%%%%%%%%%%%%%%%%%%%%%%%%%%%%%%%%%%%%%%%%%%%%%%%%%%%%%%%%%%%%%%%%%%%%%%%%%%%%

This section aims to demonstrate that time can introduce noise into the synchronization of geometric time series. Instead, by inducing the transformation in ~\eqref{eq:Spatial_sequence} thanks to the SS algorithm, one can extrapolate a skill that more accurately represents the informative content within the demonstration library. This improvement will be quantified in terms of the quality of the alignment (Sec.~\ref{subsec:Alignment Algorithms Comparisons}) and quality of the approximation (Sec.~\ref{subsec:Robot Handwriting Comparisons}). In the latter, we also evaluate the $\Delta$ optimization of Sec.~\ref{subsec:optimization}. Before that, Sec.~\ref{subsec:computational_complexity} briefly analyzes the computational demand of the SS.
Please note that all the metrics mentioned below have been briefly introduced and referenced for the sake of readability and space. Readers interested in a detailed explanation of these metrics are encouraged to consult the additional material provided in this paper.

%----------------------------------------------------------------------------
\subsection{Computational Complexity}\label{subsec:computational_complexity}
%----------------------------------------------------------------------------

%
\begin{table}[t]
\caption{Computational comparisons between the SS and other DTW algorithm versions.}
    \centering
    \begin{tabular}{ccc} 
        \hline
        & Comp. Complexity (1D) & Elapsed time [s] \\  
        \hline
        DTW\cite{muller2007dynamic}, & O$(NM)$ & $2.40\!\pm\!1.305$ \\
        cDTW\cite{sakoe1978dynamic} & O$(NW)$ & $0.82\!\pm\!0.530$ \\
        FastDTW\cite{salvador2007toward}, & O$(N)$ & $0.0080\!\pm\!0.00403$ \\
        SS ($\Delta_1\! \triangleq\! 0.001$) & O$(K_1N)$ & $0.03\!\pm\!0.014$ \\
        SS ($\Delta_2\! \triangleq\! 0.01$) & O$(K_2N)$ & $0.02\!\pm\!0.012$ \\
        \hline
    \end{tabular}
    \label{tab:comp_complexity}
\vspace{-0.4cm}    
\end{table}
\begin{figure*}[t]
\centering
    \includegraphics[trim={0cm 0cm 0cm 0cm}, width=\linewidth]{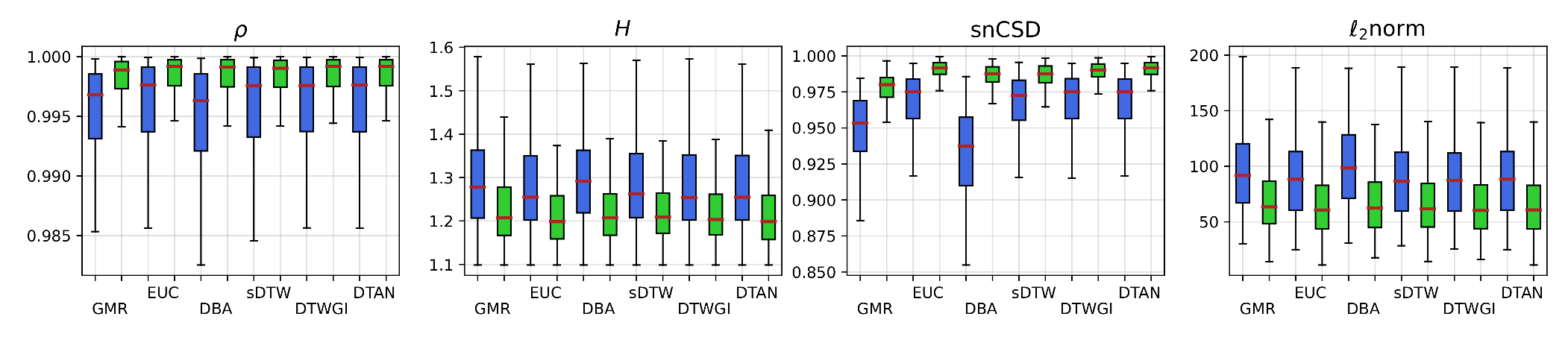}
    \caption{Box plots and table of simulation results on LASA handwriting dataset~\cite{khansari2011learning} when comparing time (blue) versus arc-length (green) parametrization obtained through SS. From the left, the box reports the following metrics: Rho ($\rho$), Entropy ($H$), Sum-Normalized Cross-Spectral Density (snCSD), $\ell_2$ norm error. }
\label{fig:box_time_arclength}
% \vspace{-0.3cm}
\end{figure*}
\begin{table*}[t]
\centering
\caption{ Mean and standard deviation values of synchrony metrics for time ($t$) versus arc-length ($s$) parameterization tests.}\vspace{-4mm} 
    \begin{center}
    \resizebox{\linewidth}{!}{
    \begin{tabular}{cccccccc} %|c|c|c|c|c|c|c|c|
     \hline
      \multicolumn{2}{c}{} & GMR & EUC & DBA & sDTW & DTWGI & DTAN \\
      \hline
      \multirow{2}{*}{$\rho$} & $\!t\!$ & $0.994\!\pm\!0.0058$ & $0.995\!\pm\!0.0061$ & $0.994\!\pm\!0.0066$ & $0.995\!\pm\!0.0061$ & $0.995\!\pm\!0.0061$ & $0.995\!\pm\!0.0060$ \\
      & \cellcolor{Gray} $\!s\!$ & \cellcolor{Gray} $0.997\!\pm\!0.0025$ & \cellcolor{Gray} $0.998\!\pm\!0.0027$ & \cellcolor{Gray}$0.998\!\pm\!0.0027$ & \cellcolor{Gray}$0.998\!\pm\!0.0026$ & \cellcolor{Gray}$0.998\!\pm\!0.0026$ & \cellcolor{Gray} $\boldsymbol{0.998\!\pm\!0.0025}$
      \\
      \hline
      \multirow{2}{*}{$H$} & $\!t\!$ & $1.292\!\pm\!0.1187$ & $1.281\!\pm\!0.1169$ & $1.301\!\pm\!0.1179$ & $1.287\!\pm\!0.1175$ & $1.281\!\pm\!0.1172$ & $1.282\!\pm\!0.1170$ \\
      & \cellcolor{Gray} $\!s\!$ &\cellcolor{Gray} $1.228\!\pm\!0.0850$ &\cellcolor{Gray} $\boldsymbol{1.214\!\pm\!0.0753}$ &\cellcolor{Gray} $1.222\!\pm\!0.0750$ & \cellcolor{Gray}$1.220\!\pm\!0.0750$ & \cellcolor{Gray}$1.218\!\pm\!0.0744$ & \cellcolor{Gray}$\boldsymbol{1.214\!\pm\!0.0753}$ \\
      \hline
      \multirow{2}{*}{snCSD} & $\!t\!$ & $0.942\!\pm\!0.0467$ & $0.963\!\pm\!0.0437$ & $0.927\!\pm\!0.0482$ & $0.961\!\pm\!0.0440$ & $0.963\!\pm\!0.0436$ & $0.963\!\pm\!0.0437$ \\
      & \cellcolor{Gray} $\!s\!$ & \cellcolor{Gray} $0.975\!\pm\!0.0167$ &\cellcolor{Gray} $\boldsymbol{0.990\!\pm\!0.0086 }$ & \cellcolor{Gray}$0.986\!\pm\!0.0106$ & \cellcolor{Gray}$0.985\!\pm\!0.0132$ & \cellcolor{Gray}$0.988\!\pm\!0.0091$ & \cellcolor{Gray} $\boldsymbol{0.990\!\pm\!0.0086}$  \\
      \hline
      \multirow{2}{*}{$\ell_2$ norm} & $\!t\!$ & $98.371\!\pm\!46.5363$ & $92.250\!\pm\!47.1034$ & $105.341\!\pm\!52.3183$ & $92.657\!\pm\!48.4249$ & $92.168\!\pm\!47.4000$ & $92.248\!\pm\!47.1020$ \\
      & \cellcolor{Gray} $\!s\!$ & \cellcolor{Gray} $70.236\!\pm\!32.6700$ &\cellcolor{Gray}  $\boldsymbol{66.967\!\pm\!33.4304}$ & \cellcolor{Gray}$69.136\!\pm\!32.7610$ &\cellcolor{Gray} $68.717\!\pm\!32.8724$ & \cellcolor{Gray}$67.632\!\pm\!33.3254$ & \cellcolor{Gray}$66.970\!\pm\!33.4248$ \\
      \hline
    
    \end{tabular}
    }
\end{center}
\label{tab:tab1}
\vspace{-4mm}
\end{table*}
%

% The SS algorithm is filters a input time-based trajectory and returns as output its related path, moving from the time-domain to the arc-length domain. DTW, instead, is a signal processing technique or, more properly, a similarity measure used to find the optimal time alignment between two sequences of different duration~\cite{muller2007dynamic, chiu2004content}. While the two algorithms are different in nature, as trajectory alignment is investigated in this paper we used without loss of consistency the DTW as a baseline to compare the computational complexity of the SS algorithm.

The SS algorithm filters an input time-based trajectory and outputs its corresponding path in the arc-length domain, effectively shifting from time-based to geometric representation. In contrast, DTW is a signal processing technique used to optimally align sequences of different durations~\cite{muller2007dynamic, chiu2004content}. Although conceptually different, DTW serves as a baseline in this study to compare the computational complexity of the SS algorithm in the context of trajectory alignment.

We assume to analyze 1D signals, as most of DTW-related literature is based on that. The considered baselines are the original DTW formulation in\cite{muller2007dynamic}, the constrained DTW (cDTW) based the Sakoe-Chiba Band\cite{sakoe1978dynamic}, and the FastDTW in\cite{salvador2007toward}. For the SS algorithm, instead, two values of $\Delta$ with difference of one order of magnitude were considered to analyze how much the choice of this parameter influences computational costs.

Given two signals $\y_1 \in \mathds{R}^{N\times1}$ and $\y_2 \in \mathds{R}^{M\times1}$ with $N > M$, Table~\ref{tab:comp_complexity} outlines the computational complexity of the algorithms. For cDTW, $W$ defines a limited sample band for computing the DTW distance matrix ($W\ll M$). If considering $\y_1$, for $\Delta$ sufficiently low the SS algorithm would have to process all $N$ samples. Moreover, given the length $\mathcal{L}$ of the curve, the choice of $\Delta$ implies that $K=\mbox{int}(\mathcal{L}/\Delta)$ spatial samples are generated. For these considerations, SS theoretically operates with complexity O$(K\!\cdot\!N)$ in the worst case scenario, i.e. when processing all $\y_1$ samples\footnote{For multi-dimensional trajectories $y_1 \in \mathds{R}^{N\times d}$ the complexity of the SS algorithm is O$(K\!\cdot\!N\!\cdot\!d)$. Note that the original trajectory $\y_1$ could be ideally obtained with $\Delta\rightarrow0$. In practice, even non-zero $\Delta$ values allow to achieve good approximations, with low computational burden, and such that $K$ is limited.}. For a more practical evaluation, we conducted tests using the trajectory dataset presented in Sec.~\ref{subsec:Robot Handwriting Comparisons} to measure the elapsed time for executing the algorithms on the same platform. As shown in the table, FastDTW achieved the best runtime performance, which is expected given that the current implementation of our algorithm is not yet parallelized. Nonetheless, two important observations can be made: (i) the choice of $\Delta$ has minimal impact on the performance of the SS algorithm, and (ii) although all the algorithms operate offline, the SS algorithm’s performance is not a limiting factor. Moreover, SS offers unique advantages over DTW, which will be highlighted in the following sections.

%-----------------------------------------------------------------------------
\subsection{Alignment Algorithms Comparisons Based on LASA
Handwriting Dataset}\label{subsec:Alignment Algorithms Comparisons}
%-----------------------------------------------------------------------------

This simulation experiment evaluates the effectiveness of the SS algorithm in improving time series alignment. To achieve this, we conducted experiments using the LASA handwriting dataset~\cite{khansari2011learning},  a well-established benchmark for diverse robotic applications. For each handwriting in the dataset, we calculated the corresponding barycenter, which served as a reference for aligning each group. Simulations were conducted both in the time domain and using the SS algorithm to move to the arc-length domain of the handwriting curves. For the SS algorithm, we choose $\Delta = 0.02$. In both domains, we calculated metrics to evaluate the level of alignment between the handwriting and their respective barycenter.

\begin{figure*}[t]
    \centering
    % First column (3 pictures)
    \begin{minipage}{0.15\textwidth}
        \centering
        \includegraphics[width=0.7\linewidth]{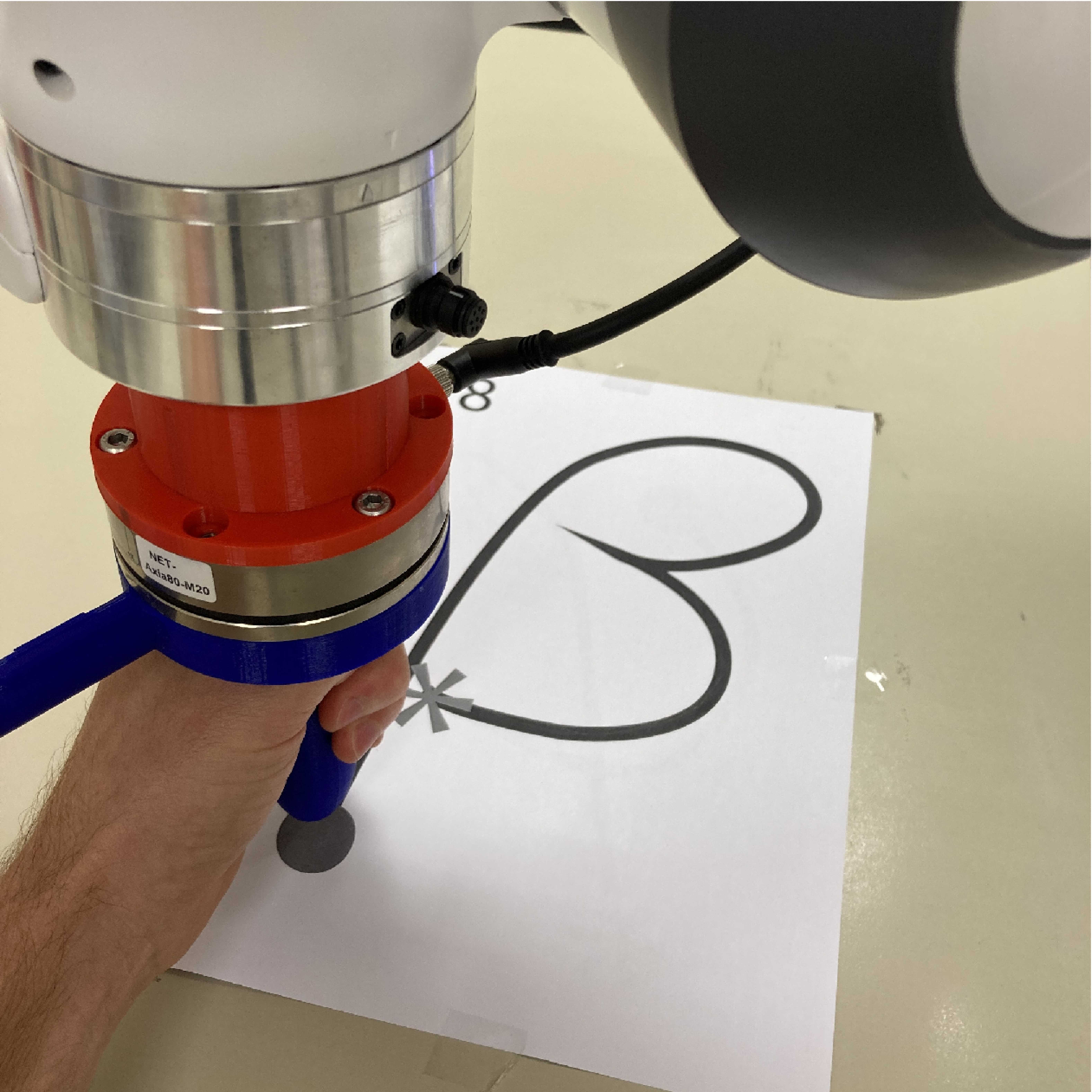}
        \caption*{}
        \includegraphics[width=0.7\linewidth]{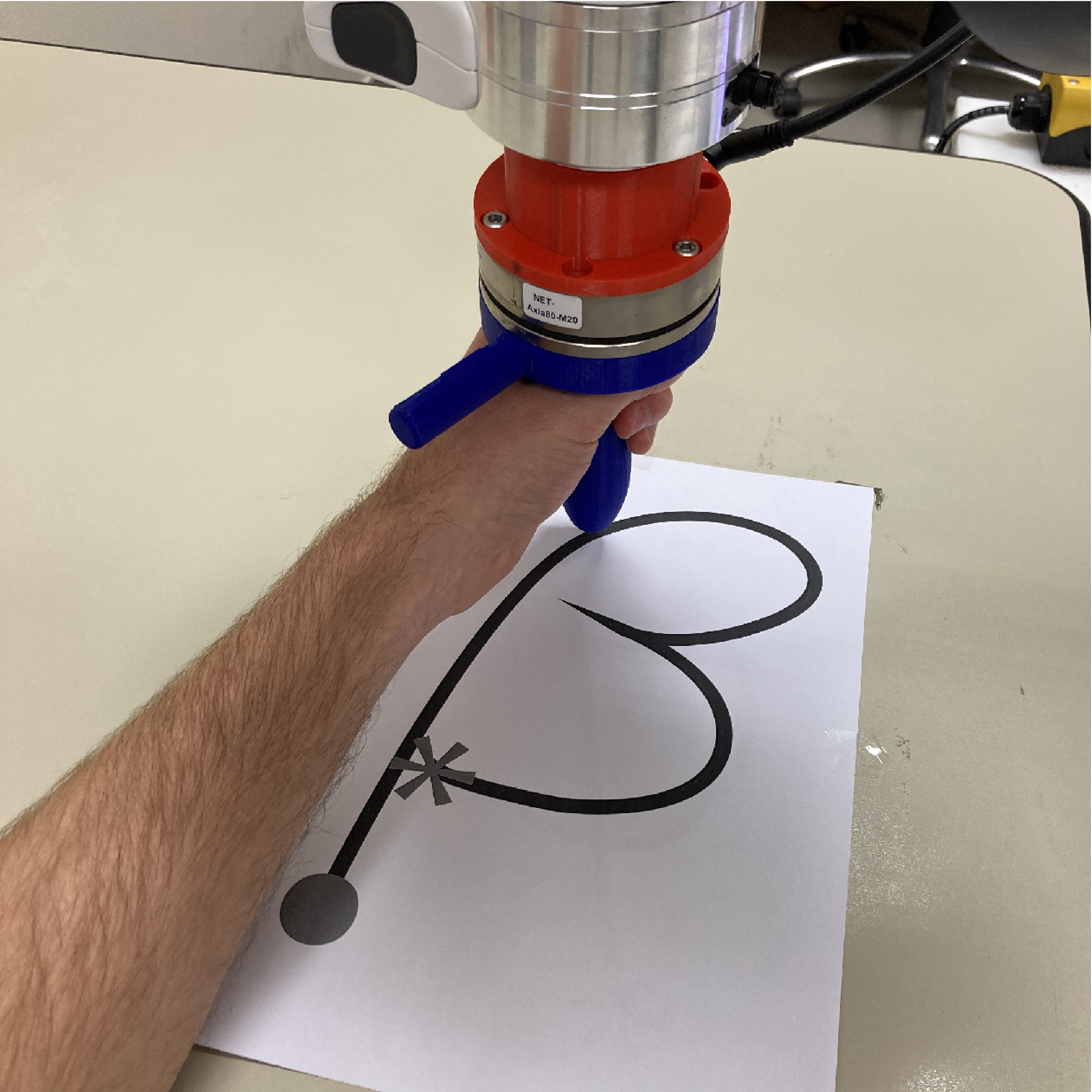}
        \caption*{}
        \includegraphics[width=0.7\linewidth]{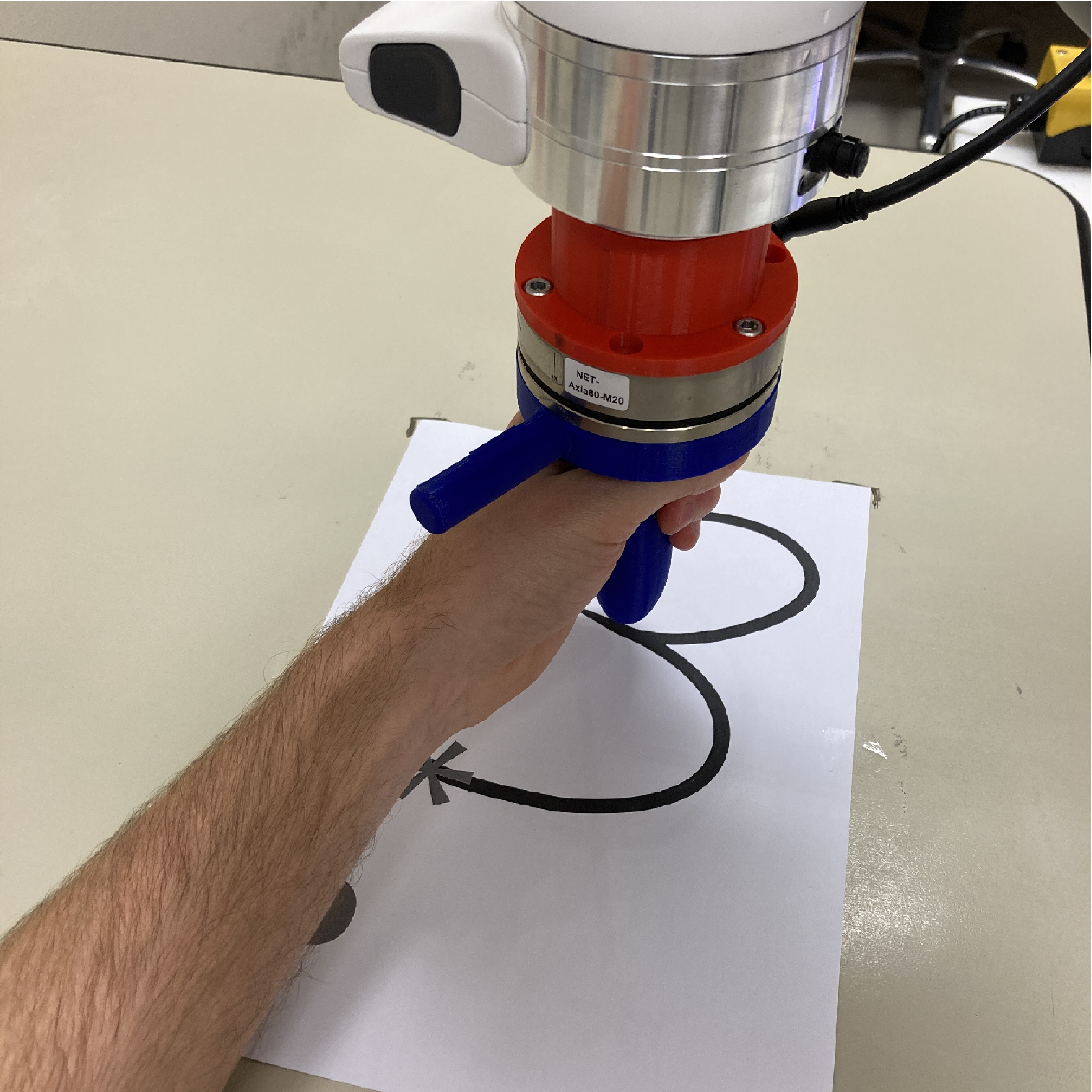}
        \caption*{(a)}
    \end{minipage}
    \hfill
    % Second column (2 pictures)
    \begin{minipage}{0.41\textwidth}
        \centering
        \includegraphics[width=\linewidth]{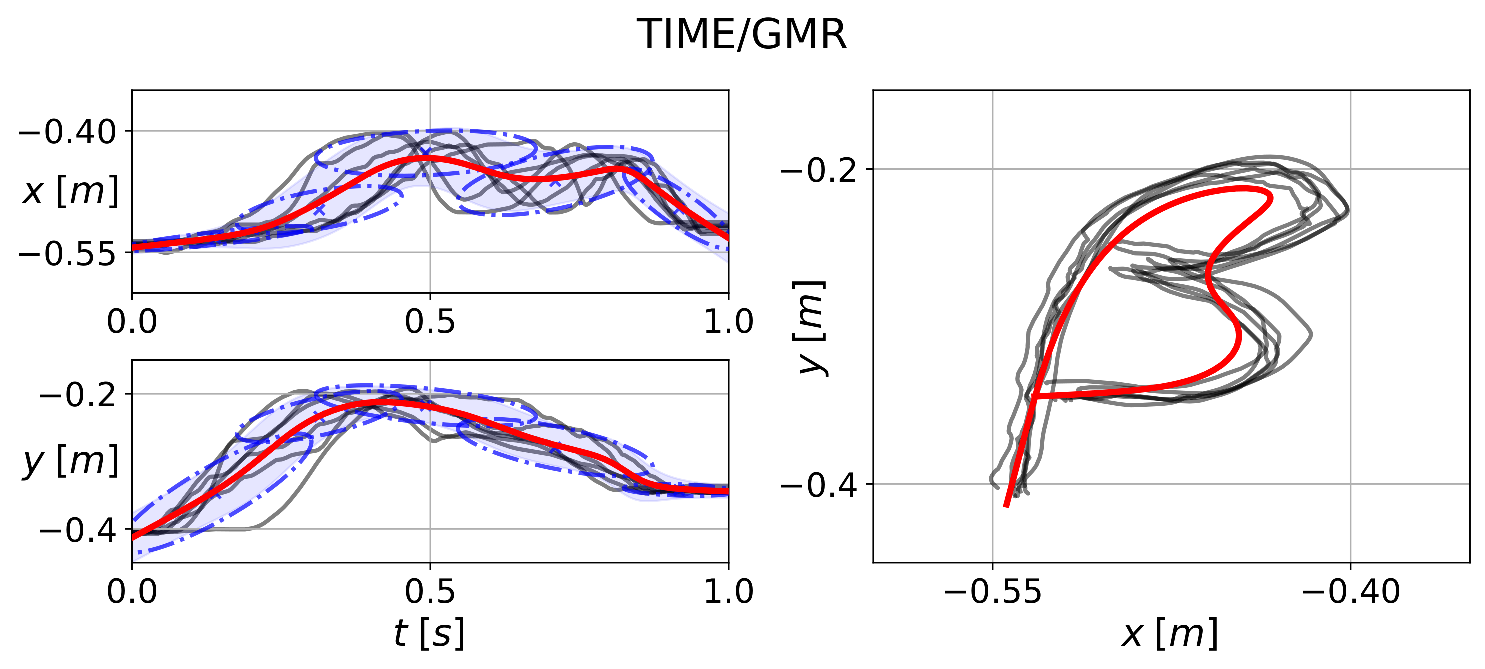}
        \caption*{(b)}
        \includegraphics[width=\linewidth]{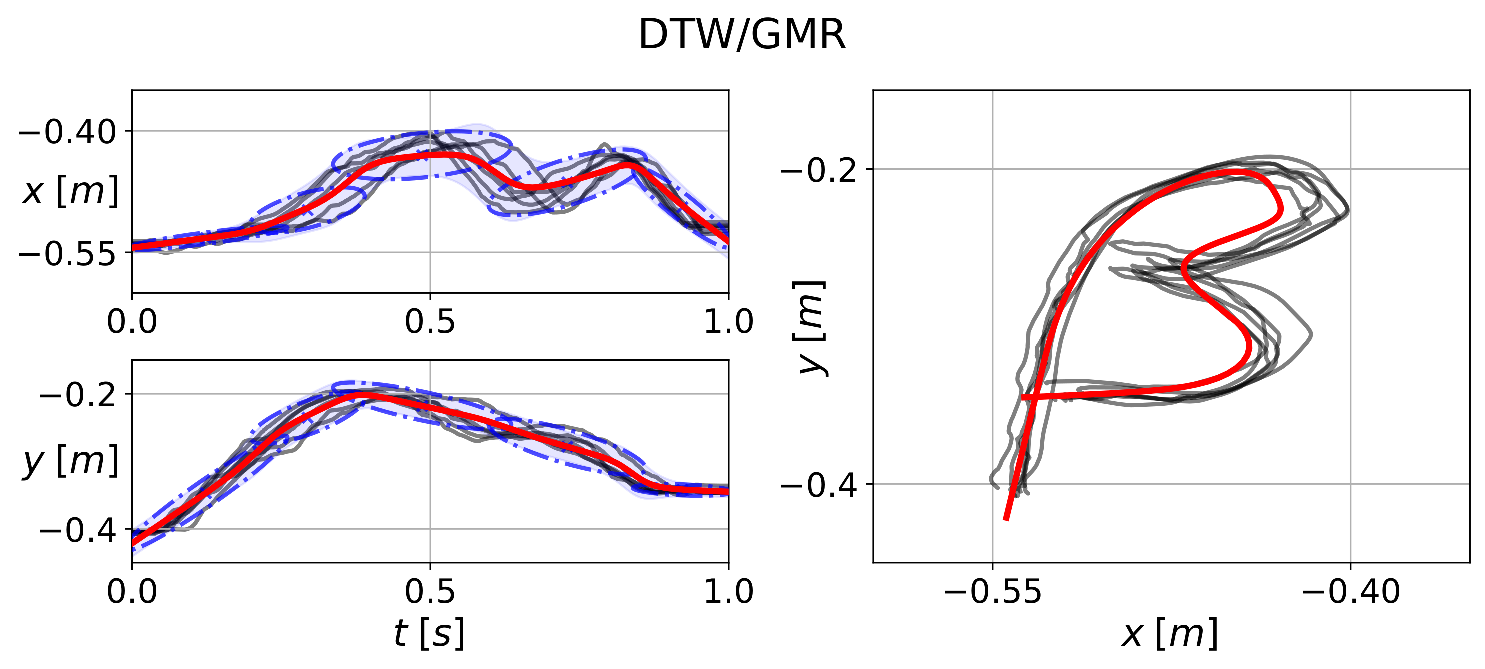}
        \caption*{(d)}
    \end{minipage}
    \hfill
    % Third column (2 pictures)
    \begin{minipage}{0.41\textwidth}
        \centering
        \includegraphics[width=\linewidth]{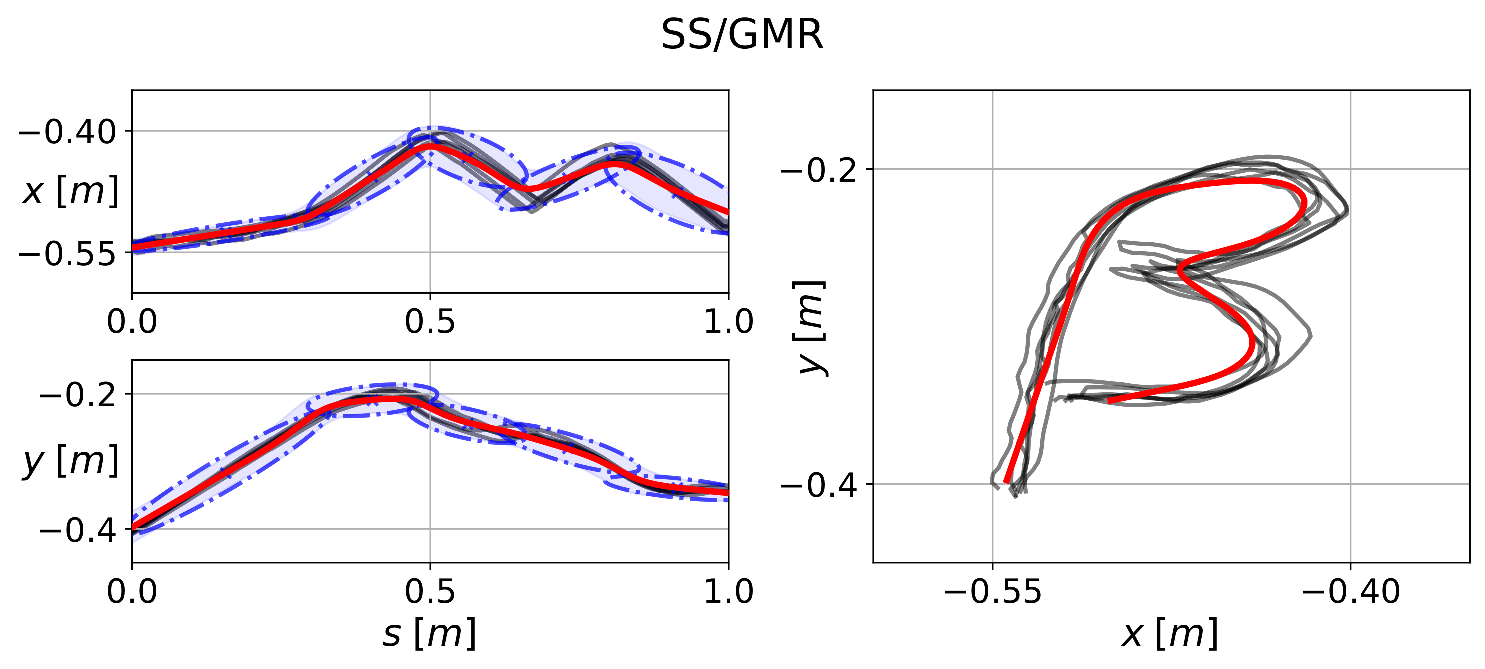}
        \caption*{(c)}
        \includegraphics[width=\linewidth]{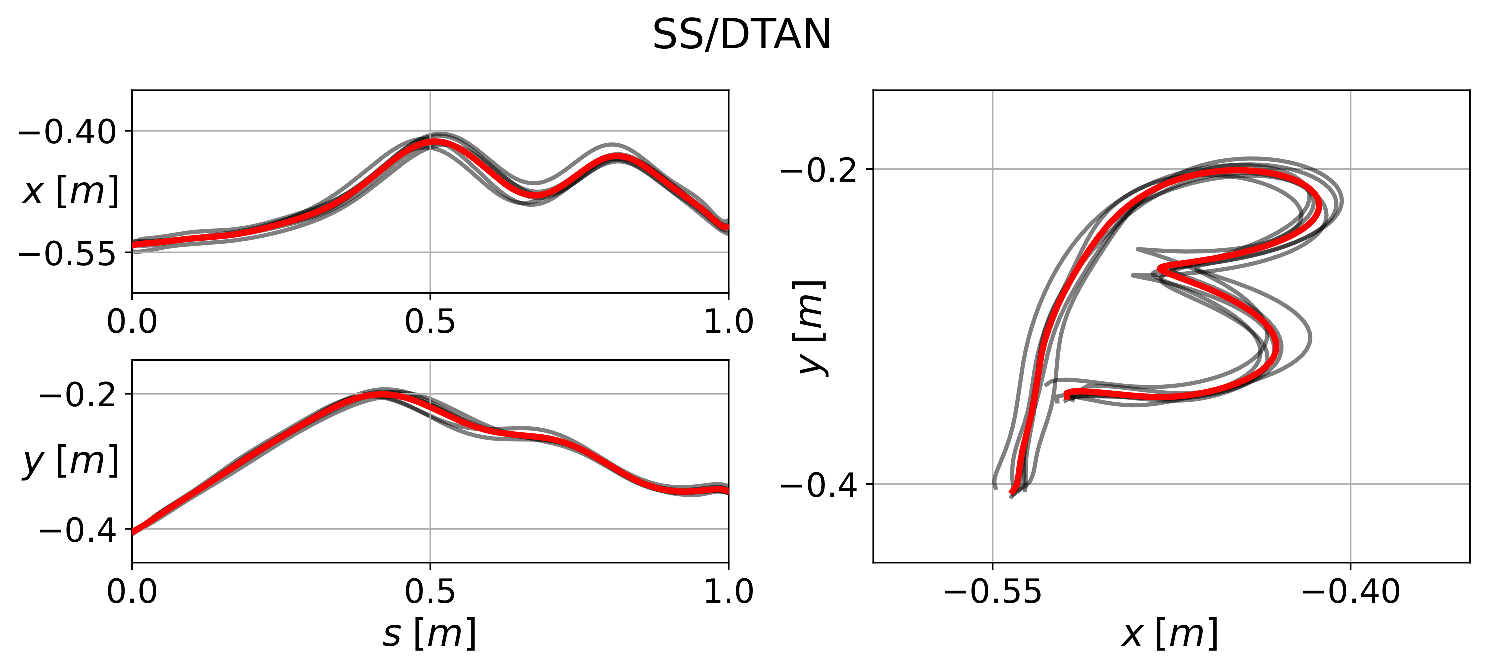}
        \caption*{(e)}
    \end{minipage}
    \caption{Frames of robot demonstrations over the AutoLab Co-Manipulation dataset (a). Barycenter computation using time-based (b)(d) and arc-length parametrized (c)(e) curves, with normalized intervals for better visualization.}
    \label{fig:geometric_experiments}
\vspace{-0.5cm}
\end{figure*}

To measure multivariate synchrony, thus not considering just pairwise relationships, we utilized the multiSyncPy package~\cite{hudson2023multisyncpy}, which comprises metrics from different information theory studies. 
For our analysis, we focused on the following metrics:
Rho ($\rho$), Entropy ($H$), Sum-Normalized Cross-Spectral Density (snCSD) and $\ell_2$ norm error, which serve the following purposes. 
% Rho
\textit{Rho ($\rho$)}~\cite{richardson2012measuring}: Rho compares the phase of signals to assess their similarity. Calculating an aggregate relative phase across multiple signals measures how closely each signal's phase aligns with the overall phase.
% Entropy
\textit{Entropy ($H$)}~\cite{stevens2014toward}: entropy quantifies the regularity and predictability of time series. Lower entropy values indicate synchronized, regular behavior, while higher entropy reflects increased variability and unpredictability.
% Sum-normalized CSD
\textit{Sum-Normalized Cross-Spectral Density (snCSD)}~\cite{hudson2023multisyncpy}: unlike coherence, which analyzes the power spectrum  of the signal and averages across frequencies, ignoring single amplitude variations~\cite{white1984signal}, snCSD accounts for the cross-spectral density at each frequency. This metric helps reduce the impact of noise, providing a less biased measure of coherence.  
% $\ell_2$ norm error
Cumulative \textit{$\ell_2$ norm error}: given two signals $\x_i$ and $\y_i$, $i=0,1,...N$ with $N$ being the number of samples, this metric computes the sum of each error $||\x_i-\y_i||_2$ and serves to express a geometric approximation metric within the aligned signals.
Since this metric requires equal-length signals, without loss of consistency, trajectories have been resampled in this section.

The results from the simulations are reported in Fig.~\ref{fig:box_time_arclength} and Table~\ref{tab:tab1}. We decided to compare different algorithms for barycenter computation, whose choice was based on (i) their popularity and (ii) the public availability of their codes. The algorithms are Gaussian Mixture Regression (GMR)~\cite{calinon2007learning}, Euclidean distance averaging (EUC)~\cite{JMLR:v21:20-091}, DTW Barycenter Averaging (DBA)~\cite{petitjean2011global}, soft-DTW (sDTW)~\cite{cuturi2017soft}, DTW with Global Invariance (DTWGI)~\cite{vayer2020time}, and Diffeomorphic Temporal Alignment Nets (DTAN)~\cite{shapira2019diffeomorphic}. 
In Fig.~\ref{fig:box_time_arclength}, the results for each algorithm are shown with the time domain in blue and the arc-length domain in green. It is clear that whenever the SS algorithm is applied, the metrics improve significantly across the dataset, as summarized by the statistics of the boxplots. This indicates synchrony is enhanced when trajectories are considered in their arc-length domain. This is crucial as it allows for extracting a more representative barycente from the demonstrations, unaffected by timing discrepancies.
In particular, Table~\ref{tab:tab1} shows that the SS-DTAN algorithm performs slightly better than other methods. Similar outcomes are seen with SS-EUC, though its performance drops when there are significant time variations in the demonstrations (e.g., pauses or varying speeds). Therefore, only DTAN will be further considered in the next section.

%-----------------------------------------------------------------------------
\subsection{Skill Synthesis from Demonstrations}\label{subsec:Robot Handwriting Comparisons}
%-----------------------------------------------------------------------------

%
\begin{figure}[t]
\centering
    \begin{subfigure}{0.95\columnwidth}    
    \setlength{\abovecaptionskip}{-2pt} 
    \centering
    \includegraphics[width=1\linewidth]{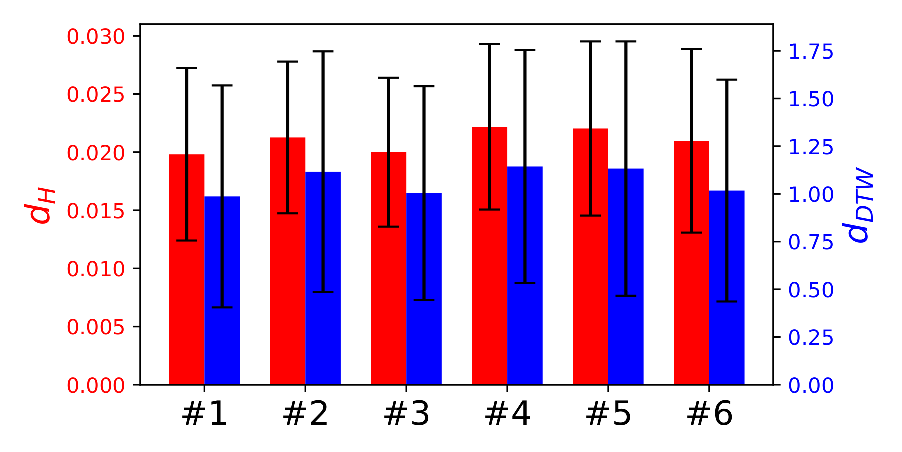}
        \caption{}
        \label{subfig:dtw_reference}
    \end{subfigure}\hspace{-5mm}
    \begin{subfigure}{0.95\columnwidth}
    \setlength{\abovecaptionskip}{-2pt} 
    \centering
        \includegraphics[width=1\linewidth]{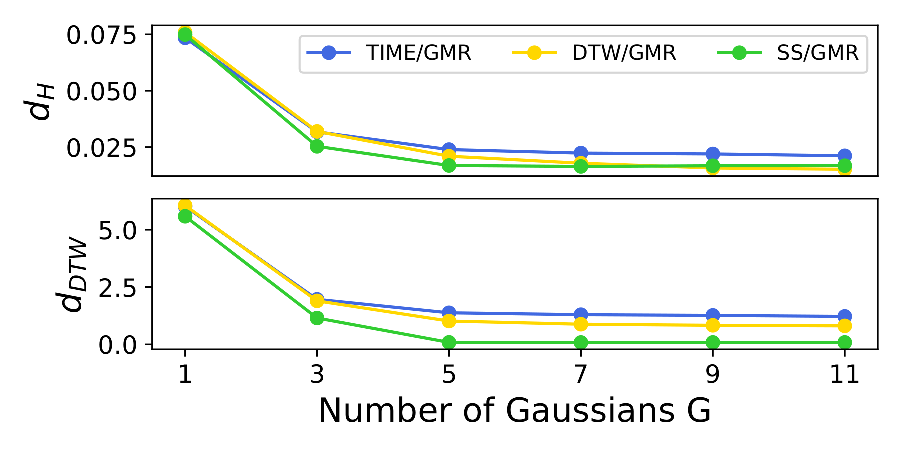}
        \caption{}
        \label{subfig:gaussians}
    \end{subfigure}
    \caption{Results of the $d_H$ and $d_{DTW}$ metrics (a) for the DTW/GMR combination across reference variation and (b), for TIME/GMR, DTW/GMR and SS/GMR when varying the number of components $G$ in the GMM model. }
    \label{fig:revisors_exp}
\vspace{-0.3cm}    
\end{figure}
This section analyzes the geometric approximation error of the computed barycenter. To do this, we created a custom dataset of end-effector position recordings.
Our dataset consists of 21 symbols recorded using a Franka Emika Panda robot in gravity compensation mode (see Fig.~\ref{fig:geometric_experiments}(a)). 
Six recordings were made for each symbol, maintaining the same start and endpoint, but varying the speed along the curve and introducing pauses at the user's discretion. 
This concern differs from the LASA handwriting dataset, where recordings have shared endpoints but varying starting positions and minimal timing discrepancies. Since LfD relies heavily on consistent demonstrations, time discrepancies might be perceived as a limitation. However, such variability is common in scenarios where the teacher faces challenges like physical constraints \cite{braglia2024phase}, and it serves to test the robustness of algorithms by introducing valuable diversity in the dataset. Furthermore, our dataset includes sharp-corner figures, which provide new research opportunities in co-manipulation tasks, where smooth paths are typically the primary focus.

% Figure \ref{fig:incremental_learning} displays how the quality of the retrieved barycenter (red) through GMR varies according to the considered number of demonstrations $N_{DIM}$ (black). While it is evident that for an higher $N_{DIM}$ one obtains a better approximation,
By using as a ground truth the computed barycenter, the approximation performances can be quantitatively evaluated using two metrics: the Hausdorff distance ($d_H$)
% ~\cite{Donoso2008}
as defined in~\eqref{Hausdorff_Distance} and, without loss of consistency, the DTW score ($d_{DTW}$)~\cite{muller2007dynamic}. 
While the former captures the maximum approximation error, the latter provides a clearer view of the overall approximation error when comparing two signals.
We tested four scenarios: (i) feeding time-based recordings to GMR (TIME/GMR), (ii) filtering recordings with SS before sending them to GMR (SS/GMR), (iii) using DTW to align trajectories before applying GMR (DTW/GMR), and (iv) combining SS with DTAN. For the SS algorithm, we choose $\Delta = 0.005$, then we observed how performances can be enhanced optimizing $\Delta$ as for Sec.~\ref{subsec:optimization}. GMR is widely used in robotics for skill extraction, which justifies its inclusion in this study. Additionally, the DTW/GMR combination has been adopted in many seminal robotics works, making it a relevant comparison algorithm~\cite{zeng2022gmmdtw}.

Figure~\ref{fig:geometric_experiments} shows the results for the $\beta$-shape recordings. The aligned demonstrated trajectories (black) were fitted with a GMM using $G=5$ components (blue) before computing the barycenter (red) with GMR, except for the SS/DTAN case where GMR was not used. In Fig.~\ref{fig:geometric_experiments}(b), temporal distortions lead to high variance, reducing the skill quality. Lower variances are observed in the SS/GMR (c) and DTW/GMR (d) cases, with the latter showing a less accurate barycenter due to the reference alignment choice. To conclude, the SS/DTAN (e) case appears to be the most promising scenario, combining the alignment provided by the SS algorithm and the regression performed by the DTAN model.

Indeed the DTW/GMR method strongly depends on selecting an appropriate reference signal for trajectory alignment, which often requires manual adjustments to achieve a satisfactory barycenter. This dependency is further demonstrated in Fig.\ref{subfig:dtw_reference}, where the metrics $d_H$ (red) and $d_{DTW}$ (blue) change with consistent variance across different reference signals. For the experiments reference \#1 was chosen. In contrast, the SS approach bypasses this issue by filtering time trajectories directly, eliminating the need for choosing an appropriate reference. Another factor to consider is the number of Gaussian components, $G$. Figure \ref{subfig:gaussians} highlights that increasing $G$ yields minimal improvements in the metrics, while it is well-known that this significantly raises computational complexity\cite{calinon2010learning}. Note that the SS/DTAN method is not considered here as it does not involve GMR.

\begin{figure}[t]
    \centering
    \begin{subfigure}{0.95\columnwidth}    
    \setlength{\abovecaptionskip}{-3pt} 
        \includegraphics[width=0.96\linewidth]{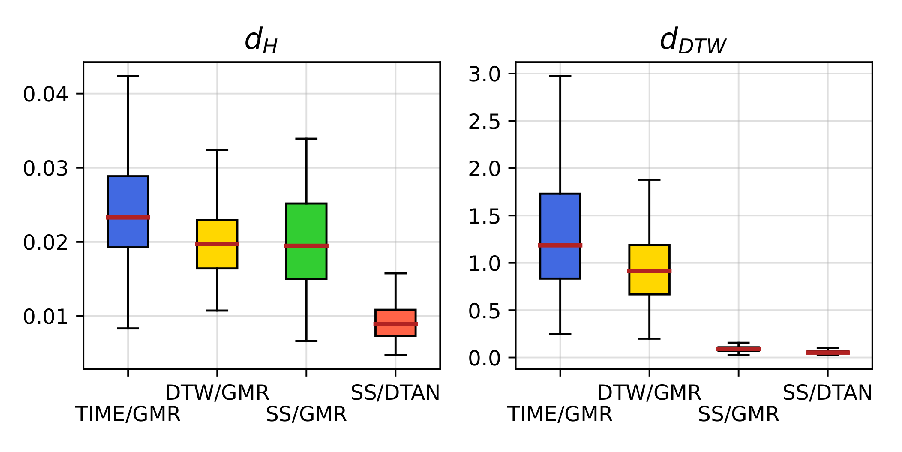} 
        \caption{}
        \label{subfig:box_distance_comparisons}
    \end{subfigure}
    \begin{subfigure}{0.95\columnwidth}  
    \setlength{\abovecaptionskip}{-3pt} 
        \includegraphics[width=0.96\linewidth]{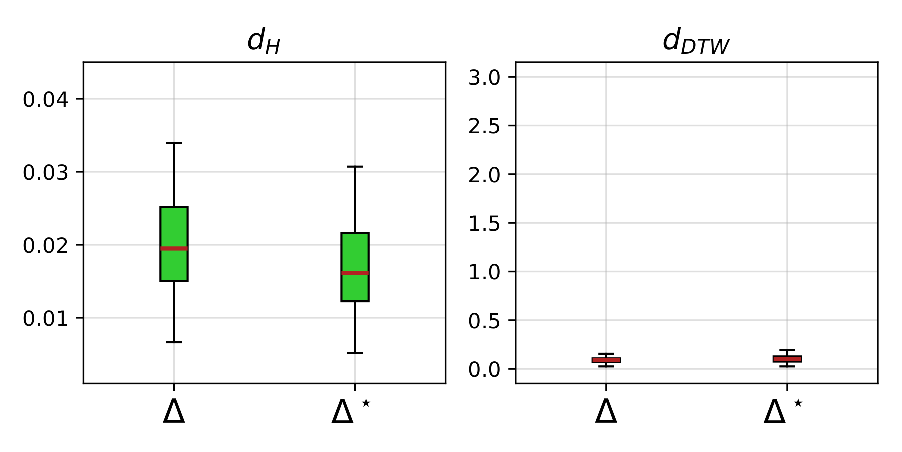} 
        \caption{}
        \label{subfig:box_optimal_delta}
    \end{subfigure}
    \caption{Results for approximation metrics experiments: (a) box plots of $d_H$ (left) and $d_{DTW}$ (right) for the compared algorithms; (b) analysis of the metrics' improvements for the SS/GMR case when computing the optimal $\Delta^\star$.}
    \label{fig:approximation_comparisons}
% \vspace{-0.5cm}    
\end{figure}

Quantitative results are presented in Fig.~\ref{subfig:box_distance_comparisons} and Table~\ref{tab:tab2} . As shown on the figure's left, the lack of prior alignment (TIME/GMR case) can result in significant displacement from the demonstrated trajectories when computing the barycenter. In contrast, both DTW/GMR and SS/GMR perform better, with comparable outcomes. However, differences become noticeable when considering the second metric, $d_{DTW}$, which evaluates the entire signal. Here, the SS/GMR combination significantly improves over both TIME/GMR and DTW/GMR. Instead, the SS/DTAN method achieves the best performance in terms of both $d_H$ and $d_{DTW}$, which can benefit from better alignment through the arc-length domain and improved barycenter computation with the DTAN algorithm. To conclude, we evaluated the SS/GMR combination by computing a desired $\Delta^\star$ for each trajectory using the algorithm in Sec.\ref{subsec:optimization}, setting $d^\star_H = 0.004$ (see Fig. \ref{subfig:box_optimal_delta}). This approach achieved a finer approximation of the dataset's trajectories, resulting in lower $d_H$ values. Although smaller $\Delta$ values increased the number of elements in the spatially-sampled trajectories, leading to higher $d_{DTW}$ values, these remained comparable to those in Table~\ref{tab:tab2}.
\begin{table}[t]
\centering
\caption{ Mean and standard deviation values for approximation metrics.}\vspace{-6mm} 
% \scriptsize
\setlength{\tabcolsep}{4pt} 
\begin{center}
        \vspace{2mm}
        \setlength{\tabcolsep}{4pt}
        \centering % Avoid extra alignment issues
        \resizebox{\linewidth}{!}{ % Resize table to fit subfigure width
        \begin{tabular}{ccccc} %|c|c|c|c|c|
        \hline
        & {\small TIME/GMR } & {\small DTW/GMR } & {\small SS/GMR } & {\small SS/DTAN } \\  
        \hline
        $d_H (e^{-\!2})$ & $2.4\!\pm\!0.81$ & $2.1\!\pm\!0.65$ & $2.0\!\pm\!0.63$ & $\boldsymbol{0.9\!\pm\!0.26}$\\
        % \hline
        $d_{DTW} $ & $1.34\!\pm\!0.645$ & $0.96\!\pm\!0.432$ & $0.09\!\pm\!0.029$ &  $\boldsymbol{ 0.06\!\pm\!0.018 }$  \\
        \hline
        \end{tabular}
        }
\label{tab:tab2}
\end{center}
\vspace{-4mm}
\end{table}
%

%-----------------------------------------------------------------------------
\subsection{Discussion}\label{subsec:Discussion}
%-----------------------------------------------------------------------------
The SS algorithm produces time-agnostic curves, changing their parametrization from time to the arc-length domain~\cite{braglia2024phase}. As shown in Section~\ref{subsec:Alignment Algorithms Comparisons}, this decoupling implicitly enhances alignment reliability and improves the performance of the analyzed algorithms.

The LASA handwriting dataset consists of trajectory recordings with a common endpoint starting from different positions, but it lacks significant timing discrepancies such as pauses or speed variations.
To further evaluate the SS algorithm, we created our own handwriting dataset. Unlike LASA, this dataset introduces significant time distortions while maintaining the same path and endpoints across repetitions. This setup allowed us to analyze the approximation error introduced when computing the barycenter of SS-filtered curves. 
We found that computing the barycenter directly from time-domain trajectories often introduces distortions. However, SS and GMR yielded performance comparable to state-of-the-art methods like DTW/GMR and a refined DTAN model due to SS's preprocessing of the trajectories. Unlike SS/GMR, DTW/GMR requires prior alignment, and DTAN needs a training phase to fine-tune parameters. In contrast, SS simply filters trajectories, balancing alignment and geometric accuracy. It also generates curves with well-defined derivatives, allowing different timing laws to be applied later, making SS a strong alternative for signal alignment in robotic trajectories.

%%%%%%%%%%%%%%%%%%%%%%%%%%%%%%%%%%%%%%%%%%%%%%%%%%%%%%%%%%%%%%%%%%%%%%%%%%%%%
\section{Conclusion}\label{sec:Conclusions}
%%%%%%%%%%%%%%%%%%%%%%%%%%%%%%%%%%%%%%%%%%%%%%%%%%%%%%%%%%%%%%%%%%%%%%%%%%%%%

In this article, we introduced a novel SS algorithm that filters input trajectories at evenly spaced points, capturing geometric information while remaining independent of timing variations. Unlike DTW, which operates in the time domain for sequence alignment, SS works in the arc-length domain, ensuring implicit alignment and time-agnostic parametrization.

We analyzed the computational complexity of SS and compared it with state-of-the-art barycenter computation algorithms, demonstrating improved alignment in the arc-length domain. Additionally, we assessed the geometric approximation error using a publicly available robot trajectory dataset. By integrating SS with GMR, we achieved more accurate barycenter approximations than traditional methods, highlighting SS as a promising approach for trajectory alignment and skill extraction. Future work will extend SS to orientation trajectories and explore applications beyond robotics.

% We address the computational complexity of SS and demonstrate its effectiveness by comparing it with several state-of-the-art barycenter computation algorithms. Outcomes show significant improvements in alignment when shifting to the arc-length domain.
% Additionally, we conducted a study to assess the geometric approximation error introduced by SS, using a publicly available dataset of robot position recordings. By combining SS with GMR for barycenter computation, we achieved more accurate approximations compared to traditional methods, highlighting SS as a promising alternative for trajectory alignment and skill extraction. In future work, we plan to extend our algorithm to handle orientation trajectories and explore its applicability beyond robotics.

\bibliographystyle{IEEEtran}

\bibliography{references}

\end{document}